\documentclass[runningheads]{llncs}
\usepackage[utf8]{inputenc}
\usepackage{graphicx}
\usepackage{amsmath,amssymb} 
\usepackage{color}

\usepackage{amsfonts,dsfont}
\usepackage[cal=boondox]{mathalfa} 
\usepackage{multirow}
\usepackage{verbatim}
\usepackage{url}

\usepackage{subcaption}
\newcommand\blfootnote[1]{%
  \begingroup
  \renewcommand\thefootnote{}\footnote{#1}%
  \addtocounter{footnote}{-1}%
  \endgroup
}

\begin{document}
\pagestyle{headings}
\mainmatter


\title{Webly Supervised Semantic Embeddings for Large Scale Zero-Shot Learning}
\titlerunning{Webly Supervised Semantic Embeddings for Large Scale ZSL}
\authorrunning{Y. Le Cacheux, A. Popescu and H. Le Borgne}

\author{Yannick Le Cacheux*\inst{1,2} \and Adrian Popescu*\inst{1} \and Hervé Le Borgne\inst{1}}
\institute{CEA LIST \and CEDRIC -- CNAM}

\maketitle


\begin{abstract}

Zero-shot learning (ZSL) makes object recognition in images possible in absence of visual training data for a part of the classes from a dataset.
When the number of classes is large, classes are usually represented by semantic class prototypes learned automatically from unannotated text collections.
This typically leads to much lower performances than with manually designed semantic prototypes such as attributes.
While most ZSL works focus on the visual aspect and reuse standard semantic prototypes learned from generic text collections, we focus on the problem of semantic class prototype design for large scale ZSL.
More specifically, we investigate the use of noisy textual metadata associated to photos as text collections, as we hypothesize they are likely to provide more plausible semantic embeddings for visual classes if exploited appropriately.
We thus make use of a source-based voting strategy to improve the robustness of semantic prototypes.
Evaluation on the large scale ImageNet dataset shows a significant improvement in ZSL performances over two strong baselines, and over usual semantic embeddings used in previous works.
We show that this improvement is obtained for several embedding methods, leading to state of the art results when one uses automatically created visual and text features.
\blfootnote{* Both authors contributed equally.}
\end{abstract}


\section{Introduction}

Zero-shot learning (ZSL) is useful when an artificial agent needs to recognize classes which have no associated visual data but can be represented by semantic knowledge~\cite{socher2013cmt}.
The agent is first trained with a set of seen classes, which have visual samples.
Then, it needs to recognize instances from either only unseen classes (classical zero-shot learning scenario) or both seen and unseen classes (generalized zero-shot learning).
To do so, it has access to visual features 
and to semantic class prototypes.
Most (generalized) zero-shot learning works focus on the proposal of adapted loss functions~\cite{frome2013devise,romera2015eszsl,akata2015sje,akata2016ale,changpinyo2016sync,annadani2018preserving} or on 
the induction of visual features for unseen classes via generative approaches~\cite{verma2017simple,bucher2018,verma2018,xian2018generating}.
Here, we use standard components for the visual part of the ZSL pipeline and instead study the influence of semantic class prototypes.
Early works exploit manually created attributes \cite{sun,cub,awa} to define these prototypes. While very efficient, such attributes require a very costly annotation effort and are difficult to scale to large datasets.
Different strategies were proposed to automate the creation of prototypes in order to tackle large scale ZSL.
An early attempt~\cite{rohrbach2011} exploited WordNet to extract part attributes.
While interesting, this method assumes that tested datasets can be mapped to WordNet, which is often impossible.
The current trend, which leverages advances in natural language processing~\cite{mikolov2013w2v,huang-etal-2012-improving,pennington-etal-2014-glove}, is to exploit standard word embeddings as semantic prototypes.
These embeddings are extracted from generic large scale text collections such as Wikipedia~\cite{huang-etal-2012-improving,mikolov2013w2v} or Common Crawl~\cite{bojanowski-etal-2017-enriching,mikolov-etal-2018-advances}.
The advantage of such methods is that prototype creation is based solely on webly supervised or unsupervised collections.
However, following~\cite{xian2018pami,hascoet19cvpr}, only standard embeddings extracted from generic collections were tested in ZSL.

We tackle the creation of semantic class prototypes for large scale ZSL via a method enabling to suitably leverage more adapted text collections for word embedding creation.
The standard generic texts are replaced by metadata associated with photo corpora because the latter are more likely to capture relevant visual relations between words.
Our method includes processing of the textual content to improve the semantic plausibility of prototypes~\cite{mikolov-etal-2018-advances} and exploits a source-based voting strategy to improve robustness of word co-occurrences~\cite{Popescu:2011,OHare2013}.
We evaluate the proposed approach for automatic building of semantic prototypes using different text collections.
We also perform an ablation study to test the robustness with respect to collection size and provide a detailed error analysis.
Results for a large scale collection show our approach enables consistent performance improvement compared to existing automatic prototypes.
Interesting performance is also obtained for smaller datasets, where the proposed prototypes reduce the gap with manual prototypes.

Our contributions can be summarized as follows:
\begin{itemize}
    \item We focus on the understudied problem of semantic prototype design for ZSL, and propose a method to create better embeddings from noisy tags datasets.
    \item We conduct extensive experiments and ablation studies to (1) demonstrate the effectiveness of the proposed method; (2) provide a variety of results with different embeddings which can be used for future fair comparison; (3) provide insight on the remaining challenges to close the gap between manual and unsupervised semantic prototypes.
    \item We collect new corpora and produce state-of-the-art semantic class prototypes for large-scale ZSL which will be released to the community. The code used to generate these prototypes will also be released.\footnote{\url{https://github.com/yannick-lc/semantic-embeddings-zsl}}
\end{itemize}


\section{Related Work}

\subsubsection{Zero-shot learning.}
Zero-shot~learning~\cite{akata2013label,lampert2009,larochelle2008,palatucci2009} attempts to classify samples belonging to \emph{unseen~classes}, for which no training samples are available. 
Visual samples are available during training for \textit{seen~classes} and both seen classes and unseen classes have ``semantic'' prototypes associated to them. 

The first ZSL approaches were introduced a decade ago~\cite{larochelle2008,lampert2009,palatucci2009} and a strong research effort has been devoted to the topic ever since~\cite{socher2013cmt,zhang2015sse,romera2015eszsl,shigeto2015lsv,rahman2018unified,norouzi2014conse,xian2016latem}. 
Several of these works relied on a triplet loss to group relevant visual sample close to the prototype in the joint space while discarding irrelevant ones~\cite{frome2013devise,akata2015sje,akata2016ale,changpinyo2016sync,changpinyo2018arxiv,annadani2018preserving}. 
In the generalized zero-shot~learning~(GZSL) setting, performance is evaluated both on seen and unseen classes~\cite{chao2016}. 
Then, a strong bias towards recognizing seen classes appears~\cite{xian2017cvpr}. 
It is nevertheless possible to tune the hyper-parameters of a ZSL method to boost its performance in a GZSL setting~\cite{lecacheux2018}. 
Recent generative approaches propose to learn discriminative models on unseen classes from artificial samples resulting from a generative model previously learned on seen classes~\cite{verma2017simple,bucher2018,verma2018,xian2018generating}.  
The transductive ZSL setting assumes that the unlabelled visual testing samples can be used during training~\cite{fu2015,kodirov2015,rohrbach2013,song2018transductive}. This usually boosts the performance, but we consider such a hypothesis too restrictive in practice, and this setting is out of the scope here.

\subsubsection{Semantic representation.}
Semantic prototypes can be created either manually or automatically.
Since the former are difficult to scale, we focus on automatically created ones, that usually rely on large-scale datasets collected on the Web. 
The extraction of word representations from the contexts in which they appear is a longstanding topic in natural language processing (NLP). 
Explicit Semantic Analysis (ESA)~\cite{Gabrilovich:2007} is an early attempt to exploit topically structured collections to derive vectorial representations of words. It proposes to represent each word by its tf-idf weights with regard to a large collection of Wikipedia entries (articles).
ESA was later improved by adding a temporal aspect to it~\cite{Radinsky:2011} or by the detection and use of concepts instead of unigrams~\cite{Hassan:2011}.
ESA and its derivates have good performance in word relatedness and text classification tasks. 
However, they are relatively difficult to scale because they live in the vectorial space defined by Wikipedia concepts which typically includes millions of entries.

The most influential word representation models in the past years are based on the exploitation of the local context.
Compared to ESA, they have the advantage of being orders of magnitude more compact, with typical sizes in the range of hundreds of dimensions.
word2vec embeddings~\cite{NIPS2013_5021} are learned from co-occurrences in local context window which are modeled using continuous bag-of-words and skip grams. This model usually outperforms bag-of-words~\cite{NIPS2013_5021,bojanowski-etal-2017-enriching,mikolov-etal-2018-advances}.
The authors of~\cite{mikolov-etal-2018-advances} analyze the role of different preprocessing steps on embedding performance. They show that the combination of tricks such as removal of duplicate sentences, phrase detection to replace unigrams, use of subword information or frequent word subsampling is beneficial.
One shortcoming of word embeddings as proposed in~\cite{NIPS2013_5021} is that they only take into account the local context of words. 
GloVe~\cite{pennington-etal-2014-glove} was introduced as an alternative method which also includes a global component obtained via matrix factorization. 
The model trains efficiently only on non-zero word-word co-occurrence matrix instead of a sparse matrix or on local windows. 
It provides superior performance compared to continuous bag-of-words and skip gram models on a series of NLP tasks, including word analogy and similarity.
The FastText model~\cite{bojanowski-etal-2017-enriching} derives from that proposed by Mikolov but considers a set of n-grams that can compose the words, compute some embeddings then represent a word as the sum of the vector representation of its n-grams. It thus models the internal structure of the words and thus allows to compute representations of out of vocabulary words. 
The state of the art in a large array of natural language processing task was recently improved by the introduction of contextual models such as ELMo~\cite{peters-etal-2018-deep}, GPT~\cite{radford-2018} or BERT~\cite{devlin-etal-2019-bert}.
These approaches make use of deep networks and model language at sentence level instead of word level as was the case for skip grams and GloVe.
While very interesting for tasks in which words are contextualized, they are not directly applicable to our ZSL scenario which requires the representation of individual words/class names. 

\subsubsection{Multimodal representations.}
The word representation approaches presented above exploit only textual resources and there are also attempts to create multimodal word embeddings. 
For instance, vis-w2v~\cite{Kottur_2016_CVPR} exploits synthetic scenes to learn visual relations between classes. 
The main challenge here is to model the diversity of natural scenes via synthetic scenes.
ViCo~\cite{gupta2019vico} exploit word co-occurrences in natural images in order to improve purely textual embeddings like GloVe. 
They show that visual and textual components complement each other well and provide SotA performance in tasks such as visual question answering, image retrieval or image captioning.
However, ViCo is not usable in ZSL because it can only improve the representation of a word if some images of it are available.
This drawback is inherent to all multimodal word representations and we thus focus on improving purely textual representations. Regarding visual features only, \cite{joulin16learning,vo17cviu} showed that one can train convolutional networks on a dataset of unannotated images collected on the Web, and that these networks perform well in a transfer learning context. 

Previous works in ZSL used embeddings to represent the semantic prototype, either at a small scale on CUB~\cite{akata16multi_cue} or at a larger scale on ImageNet, using word2vec~\cite{frome2013devise,chao2016,lecacheux19iccv} (possibly trained on wikipedia~\cite{changpinyo2016sync,xian2018pami}) or GloVe~\cite{changpinyo2018arxiv,hascoet19cvpr}. However, they only use publicly available pre-trained models, while we propose a method to design prototypes that perform better in a ZSL context.


\section{Semantic Class Prototypes for Large Scale ZSL}

\subsubsection{Problem formulation.}
The zero-shot learning (ZSL) task considers a set $\mathcal{C}_s$ of \emph{seen} classes used during training and a set $\mathcal{C}_u$ of \emph{unseen} classes that are available for the test only. In generalized zero-shot learning (GZSL), additional samples from the seen classes are used for testing as well. However, in both cases, $\mathcal{C}_s \cap \mathcal{C}_u = \emptyset$. Each class has a semantic \emph{class prototype} $\mathbf{s}_c\in\mathbb{R}^K$ that characterizes it. 
We consider a training set $\{(\mathbf{x}_i,y_i), i=1\dots N\}$ with labels $y_i\in\mathcal{C}_s$ and visual features $\mathbf{x}_i\in\mathbb{R}^D$. The task is to learn a compatibility function $f: \mathbb{R}^D \times \mathbb{R}^K \rightarrow \mathbb{R}$ assigning a similarity score to a visual sample $\mathbf{x}$ and a class prototype $\mathbf{s}$. $f$ is usually obtained by minimizing a regularized loss function:
\begin{equation} \label{eq:objective}
\frac{1}{N} \sum_{i=1}^{N} \sum_{c=1}^{|\mathcal{C}_s|} \mathcal{L}(f(\mathbf{x}_i, \mathbf{s}_c), y_i) + \lambda\Omega[f]
\end{equation}
where $\Omega$ is a regularization term weighted by $\lambda$ which constrains the parameters of $f$, and $\mathcal{L}$ is a loss function. Once a function $f$ is learned, the testing phase consists in determining the label $\hat{y}\in\mathcal{C}_u$ (or $\hat{y}\in\mathcal{C}_s \cup \mathcal{C}_u$ for GZSL) corresponding to a visual sample $\mathbf{x}$ such that 
$\hat{y} = \underset{c \in \mathcal{C}_u}{\text{arg\,max }} f(\mathbf{x}, \mathbf{s}_c)$.

We propose to automatically derive semantic class prototypes $\mathbf{s}_c$ with a method able to adequately leverage noisy corpora which are adapted for visual tasks instead of standard text corpora previously used in ZSL~\cite{mikolov2013w2v,pennington-etal-2014-glove,bojanowski-etal-2017-enriching}.
More specifically, a corpus must contain enough visual information to enable to learn discriminative embeddings.  
We therefore create two corpora, $\mathbf{fl_{wiki}}$ and $\mathbf{fl_{cust}}$, with this goal in mind. 

\subsubsection{Corpus collection.}
$\mathbf{fl_{wiki}}$ is constituted based on Wikipedia.
We select salient concepts by ranking English Wikipedia entries by their number of incoming links and keeping the top $120,000$ of the list. 
The default Flickr ranking algorithm is then used to collect up to 5000 photo metadata for each concept.
Metadata fields which are exploited here include: (1) \textit{title} - a free text description of the photo (2) \textit{tags} - a list of tags attributed to the photo and (3) the unique identifier of the user.
Note that there is no guarantee as to the relevance of textual metadata for the content of each photo since the users are free to upload any text they wish. 
Also, photo annotations can be made in any language.
We illustrate title and tags from Flickr with the following examples:\textit{``Ísmáfur Pagophila eburnea Ivory Gull''} and \textit{``minnesota flying inflight gull arctic juvenile duluth rare lakesuperior canalpark ivorygull saintlouiscounty''}.
The title includes the Icelandic, Latin and English variants of the name while the tags give indications about the location and activity of the ivory gull.
Importantly, tags can be single words (\textit{``gull''}) or concatenated ones  (\textit{``ivorygull''},\textit{``lakesuperior''}).
This first collection is made of 62.7 million image metadata pieces and 1.11 billion words.

The ${fl_{wiki}}$ collection allows to learn generic embeddings that can be used to address large scale ZSL. However, these embeddings are still quite ``generic'' since they are representative of the Wikipedia concepts. 
For a given ZSL problem, the visual samples of unseen classes are unknown during training, but the name of these classes can be known before the actual production (testing) phase. 
Such a hypothesis is implicitly made by most generative ZSL approaches, which synthesize faked visual samples from the prototype only~\cite{verma2017simple,bucher2018,verma2018,xian2018generating}.
Following a similar hypothesis, we build $\mathbf{fl_{cust}}$, a custom subset of FlickR, which is built using the class names from the three ZSL used in evaluation datasets (ImageNet-ZSL, CUB and AWA). 
The collection process is similar to that deployed for $fl_{wiki}$.
The only difference is that we use specific class names, which may each have several variants.
This collection includes 61.9 million metadata pieces and 995 million unique words.

Each collection therefore consists in a list of $C \leq 120,000$ concepts. For each class $c$, we have a metadata set $\mathcal{M}_c = \{\mathcal{m}_1, \dots, \mathcal{m}_{N_c}\}$ made of $N_c \leq 5,000$ metadata pieces. Each metadata piece $\mathcal{m}_n$ consists in a user ID $id_n$ and a list of $T_n$ words $\mathcal{W}_n = \{w_1, \dots, w_{T_n}\}$, where the words are extracted from titles and tags. $T_n$ is typically in the range of one to two dozens.
Note that stop words were discarded during preprocessing.

\subsubsection{Creation of embeddings.}
To create text representations, a vocabulary $\mathcal{V} = \{v_1, \dots, v_V\}$ is constituted to include all $V$ distinct words in the corpus. We similarly create a set $\mathcal{U} = \{u_1, \dots, u_U\}$ of all distinct users IDs.
The usual skip-gram task~\cite{mikolov2013w2v} aims to find word representations which contain predictive information regarding the words surrounding a given word. Given sequences $\{w_1, \dots, w_T\}$ of $T$ training words such that $w_t \in \mathcal{V}$ and a context of size $S$, the objective is to maximize
\begin{equation} \label{eq:skipgram_objective}
\sum_{c,n,t=1}^{C,N_c,T_n}
\sum_{\substack{-S\leq i\leq S \\ i \neq 0}} \text{log }p(w_{t+i}|w_t)
\end{equation}

Writing $v_{w_t} \in \mathcal{V}$ the unique word associated with the $t$\textsuperscript{th} training word $w_t$ and $\mathbf{v}_{w_t}$ and $\mathbf{v}_{w_t}'$ the corresponding ``input'' and ``output'' vector representations, $p(w_i|w_t)$ is computed such that
\begin{equation}
p(w_i|w_t) = \frac{\text{exp}(\mathbf{v}_{w_i}^\top\mathbf{v}_{w_t}')}{\sum_{j=1}^V \text{exp}(\mathbf{v}_j^\top\mathbf{v}_{w_t}')}
\end{equation}

Unlike in standard text collections, such as Wikipedia, the order of words in each metadata collection $\mathcal{M}_n$ is arbitrary.
Consequently, using a fixed size window to capture the context of a word is not suitable.
We tested the use of fixed size windows in preliminary experiments and results were suboptimal.

Instead, we form all distinct word pairs $(v_{i},v_{j}), ~i \neq j$, with $v_{i}, v_{j} \in \mathcal{W}_n$, for each piece of content associated to class $c$ and feed them as training examples to the word embedding algorithms.
Pairs extracted from all concept-related metadata collections $\mathcal{M}_c = \{\mathcal{m}_n\}_{n=1}^{N_c}$ associated with concept $c$ are concatenated to form the training dataset of words embeddings, so that the objective becomes:

\begin{equation}
\sum_{c=1}^C
\sum_{\substack{(v_i, v_j),\, i \neq j \\ v_i, v_j \in \{\mathcal{m}_n\}_{n=1}^{N_c}}}
\text{log }p(v_i|v_j)
\end{equation}


\subsubsection{Addressing repetitive tags.}
It is noteworthy that many users perform bulk tagging~\cite{OHare2013} which consists in attributing the same textual description to a whole set of photos.
Users also do semi-bulk, i.e. they attribute a part of tags to an entire photo set and then complete these annotations with photo-specific tags.
Bulk is known to bias language models obtained from Flickr~\cite{OHare2013,Popescu:2011} and we propose a simple but efficient way to remove it in the next section.
To account for the bulk tagging problem, we add an additional processing step for the two collections.
The authors of~\cite{Popescu:2011} and~\cite{OHare2013} suggested to replace simple tag co-occurrences by the number of distinct Flickr users who associated the two words and reported interesting gains in image retrieval and automatic geotagging respectively.

We consequently select unique triplets $(v_{i},v_{j},u_n)$ from $\mathcal{M}_c$ for each concept $c$, so that training objective becomes
\begin{equation}
\sum_{c=1}^C ~
\sum_{\substack{(v_i, v_j, u_k),\,i \neq j \\ v_i, v_j, u_k \in \mathcal{M}_c}}
\text{log }p(v_i|v_j)
\end{equation}
This translates into adding a pair $(v_{i},v_{j})$ in the training file only once for each user and thus avoiding the effect of bulk tagging.
A positive side effect of filtering pairs with unique users is that the size of the training file is reduced and embeddings are learned faster.
A comparison of performance obtained with raw co-occurrence and with user voting is provided in the supplementary material.

The same ideas can easily be applied to other word embedding approaches. In the next section, we provide experimental results with three such approaches: word2vec~\cite{mikolov2013w2v}, GloVe~\cite{pennington-etal-2014-glove} and FastText\cite{bojanowski-etal-2017-enriching}.


\section{Experiments}

\subsection{Evaluation protocol}

\subsubsection{Baseline methods.}
To the best of our knowledge, our work is the first to explicitly address the problem of semantic class prototype design for large scale ZSL. We compare to the pre-trained embeddings (noted \textbf{pt}), as they are usually used in previous ZSL works~\cite{frome2013devise,hascoet19cvpr,lecacheux19iccv}. word2vec is trained on Google News with 100 billion words, GloVe is trained on Common Crawl with 840 billion words and the same collection with 600 billion words is used for FastText. 

We also propose two baseline methods, (\textbf{wiki}) and (\textbf{clue}), to which ours can be fairly compared. They consist in learning the embeddings from two different text collections. 
Wikipedia (\textbf{wiki}) is classically exploited to create embeddings because it covers a wide array of topics~\cite{Gabrilovich:2007}.
\textit{wiki} content is made of entries which describe unambiguous concepts with well formed sentences such as \textit{``The ivory gull is found in the Arctic, in the northernmost parts of Europe and North America.''}.
The encyclopedia provides good baseline models for a wide variety of tasks~\cite{mikolov2013w2v,mikolov-etal-2018-advances,pennington-etal-2014-glove}.
Here we exploit a dump from January 2019 which includes 20.84 billion words. It is the same data as that from which were extracted the $120,000$ concepts for our method. While useful to create transferable embeddings, Wikipedia text does nevertheless not specifically describe visual relations between words. The second baseline is based on visually oriented textual content similar to the one used in our method. The ClueWeb12~\cite{clueweb12} collection (\textbf{clue}) consists of over 700 million Web pages which were collected so as to cover a wide variety of topics and to avoid spam.
We extracted visual metadata from the \textit{title} and \textit{alt} HTML attributes associated to $clue$ images. The title content is quite similar to that we extracted from FlickR in our method. 
\textit{clue} content is often made of short texts such as \textit{``ivory gull flying''} which does not encode a lot of context.
After sentence deduplication~\cite{mikolov-etal-2018-advances}, the resulting collection includes 628 million unique metadata pieces and 3.69 billion words.

\subsubsection{Evaluation datasets.}
The generic object recognition in ZSL requires to be evaluated at a large scale and is thus usually conducted on ImageNet~\cite{imagenet}. Frome et al.~\cite{frome2013devise} proposed to use the $1,000$ classes of ILSVRC for training and different subsets of the remaining $20,841$ classes to test. 
However, it has been recently showed that a structural bias appears in this setting which allows a ``trivial model'' to outperform most existing ZSL models~\cite{hascoet19cvpr}.
For this reason, we adopt the evaluation protocol proposed by Hascoet \textit{et al.} that considers the same training classes as Frome \textit{et al.} but uses $500$ classes with a minimal structural bias for testing~\cite{hascoet19cvpr}.

To get insight into the gap existing between manual attributes and unsupervised embeddings, we also conduct experiments on two smaller benchmarks on which the ZSL task is usually conducted with manual attributes specific to each dataset: Caltech UCSD Birds 200-2011 (CUB)~\cite{cub} and Animals with Attributes 2 (AwA2)~\cite{xian2018pami}. CUB is a fine-grained dataset of 11788 pictures representing 200 bird species and AWA2 a coarse-grained dataset of 37322 pictures depicting 50 animal species.
The manual attributes of CUB and AwA2 are respectively 312 and 85-dimensional.
In our setting, we are only concerned with semantic prototypes which can be obtained automatically; our results therefore cannot be directly compared to the state-of-the-art algorithms which exploit manual attributes.
For CUB and AWA2, we adopt the experimental protocol of Xian \textit{et al.}~\cite{xian2018pami} which relies on \textit{proposed splits} that avoid any overlap between the (unseen) test classes and the ImageNet classes used to pretrain visual features on ILSVRC. For ImageNet, we use the same visual features as~\cite{hascoet19cvpr} while for CUB and AwA2 we adopt those of~\cite{xian2018pami}.

\subsubsection{ZSL methods.}
Experiments are conducted with different existing ZSL methods: we provide results for DeViSE~\cite{frome2013devise}, ESZSL~\cite{romera2015eszsl} and ConSE~\cite{norouzi2014conse} as they are the three standard methods used in \cite{hascoet19cvpr}, and therefore the only methods for which comparable results are currently available.
Although results for other models -- namely GCN-6~\cite{wang2018zero}, GCN-2 and ADGPM~\cite{kampffmeyer2019rethinking} -- are also reported in \cite{hascoet19cvpr}, these models are based on graph-convolutional networks~\cite{kipf2016semi} which make use of additional intermediate nodes in the WordNet hierarchy. Such methods are outside the scope of this study.
We additionally provide results for SynC~\cite{changpinyo2016sync} as well as two linear methods, consisting in a linear projection from the visual to the visual space ($\text{Linear}_{V \rightarrow S}$),
and a linear projection from the semantic to the visual space ($\text{Linear}_{S \rightarrow V}$) inspired by~\cite{shigeto2015lsv}, who proposed to compute similarities in the visual space to avoid the hubness problem~\cite{radovanovic}.

We train the models with the usual protocol for ZSL: hyperparameters are determined using a subset of training classes as validation. We sample respectively 200 and 50 such classes at random among the 1000 and 150 training classes of ImageNet and CUB, and use the 8 classes not in ILSVRC among the 40 training classes of AwA2.
Since ConSE and DeViSE results depend on a random initialization of the models' parameters, we report results averaged over 5 runs for these two models.

\subsubsection{Implementation details.}
Word embeddings are computed using the original implementations of word2vec~\cite{mikolov2013w2v}, GloVe~\cite{pennington-etal-2014-glove} and FastText\cite{bojanowski-etal-2017-enriching}, with the same hyperparameters (see supplementary materials). In particular, we follow the usual text processing steps they propose.
Semantic prototypes for all classes are computed using the same protocol as \cite{hascoet19cvpr} for fair comparison.
For the same reason, we use the implementation from \cite{hascoet19cvpr} to run DeViSE, ESZSL, ConSE. We use the implementation from \cite{changpinyo2016sync} for SynC, and use a custom straightforward implementation for $\text{Linear}_{V \rightarrow S}$ and $\text{Linear}_{S \rightarrow V}$.
All semantic prototypes are $\ell 2$-normalized except with ESZSL to have a setting similar to \cite{hascoet19cvpr} when applicable. We report results without such a normalization in the supplementary materials, even though the trend is mostly the same.

\begin{table*}[tb]
\caption{ZSL accuracy at large scale (ImageNet dataset), for three embedding models. Each time, the three baselines ($pt$, $wiki$ and $clue$) are compared to our method $\mathbf{fl_{wiki}}$ and its variation $\mathbf{fl_{cust}}$. 
Results marked with ``*'' correspond to a setting close to Table~2 from Hascoet \textit{et al.}~\cite{hascoet19cvpr}, and are consistent with reported results.
}
\label{tab:imagenet}
\begin{center}
\resizebox{\textwidth}{!}{
\begin{tabular}{|l|l|l|l|l|l|l|l|l|l|l|l|l|l|l|l|}
\hline
Model &  \multicolumn{5}{c|}{word2vec} &  \multicolumn{5}{c|}{GloVe}&  \multicolumn{5}{c|}{FastText}
\\ \hline
Source & \textit{pt} & wiki & clue & $\mathbf{fl_{wiki}}$  & $\mathbf{fl_{cust}}$ & \textit{pt} & wiki & clue & $\mathbf{fl_{wiki}}$  & $\mathbf{fl_{cust}}$ & \textit{pt} & wiki & clue & $\mathbf{fl_{wiki}}$  & $\mathbf{fl_{cust}}$
\\ \hline
$\text{Linear}^{}_{V \rightarrow S}$ & \textit{6.8} & 9.8 & 9.6 & 10.5 & 12.6 & \textit{10.2} & 6.2 & 4.2 & 9.6 & 9.2 & \textit{6.0} & 8.9 & 2.8 & 11.6 & \textbf{14.2} 
\\ \hline
$\text{Linear}^{}_{S \rightarrow V}$ & \textit{11.6} & 11.8 & 12.2 & 12.8 & 17.1 & \textit{14.1} & 7.9 & 8.0 & 9.2 & 11.4  & \textit{14.4} & 12.1 & 8.0  & 13.3 & \textbf{17.2} 
\\ \hline 
ESZSL & \textit{10.5}  & 10.0 & 10.7 & 9.5 & 15.3 & \textit{14.1*} & 8.0 & 10.3 & 11.1 & 12.0 & \textit{14.2} & 10.1 & 1.1 & 11.9 & \textbf{15.8} 
\\ \hline
$\text{ConSE}^{}$  & \textit{9.9} & 10.5 & 11.3 & 11.9 & 13.5 & \textit{11.3*} & 8.1 & 7.8 & 11.3 & 11.9 & \textit{11.0} & 10.5 & 5.4 & 12.6 & \textbf{14.5} 
\\ \hline
$\text{Devise}^{}$  & \textit{9.0} & 9.8 & 9.9 & 9.6 & 13.3 & \textit{11.0*} & 5.9 & 5.4 & 3.8 & 3.4 & \textit{12.3} & 10.1 & 5.6  & 10.3 & \textbf{13.8} 
\\ \hline
$\text{SynC}_{o-vs-o}^{}$ & \textit{12.2} & 12.4 & 12.6  & 12.5 & 16.3 & \textit{15.0} & 10.9 & 11.2 & 12.4 & 13.3 & \textit{14.6} & 12.6 & 7.0 & 13.2 & \textbf{16.5} 
\\ \hline
\end{tabular}
}
\end{center}
\end{table*}

\begin{table*}[bt]
\caption{ZSL accuracy at smaller scale with unsupervised semantic class prototypes. Results are reported on the CUB and Awa2 datasets, for three embedding models.
}
\label{tab:cub_awa2}
\begin{center}
\resizebox{\textwidth}{!}{
\begin{tabular}{|l|l|l|l|l|l|l|l|l|l|l|l|l|l|l|l|}
\hline
Model &  \multicolumn{5}{c|}{word2vec} &  \multicolumn{5}{c|}{GloVe}&  \multicolumn{5}{c|}{FastText}
\\ \hline
Source & \textit{pt} & wiki & clue & $\mathbf{fl_{wiki}}$  & $\mathbf{fl_{cust}}$ & \textit{pt } & wiki & clue & $\mathbf{fl_{wiki}}$  & $\mathbf{fl_{cust}}$ & \textit{pt} & wiki & clue & $\mathbf{fl_{wiki}}$  & $\mathbf{fl_{cust}}$
\\ \hline

\multicolumn{16}{|c|}{CUB dataset} \\
\hline
$\text{Linear}^{}_{V \rightarrow S}$ & \textit{7.5} & 14.0 & 13.9 & 12.2 & 16.3 & \textit{8.0} & 11.6 & 9.8 & 12.7 & 14.2 & \textit{7.2} & 13.8 & 12.2 & 11.6 & \textbf{17.5}
\\ \hline 
$\text{Linear}^{}_{S \rightarrow V}$ & \textit{11.3}  & 18.0 & 17.2 & 21.5 & 23.0 & \textit{18.2} & 16.0 & 13.4 & 14.6 & 19.0 & \textit{ 16.1} & 16.2 & 16.0 & 19.9 & \textbf{24.4}
\\ \hline 
ESZSL & \textit{15.8}  & 20.4 &  17.9 & 23.0 & 25.2 & \textit{19.9} & 17.5 & 16.9 & 19.0 & 20.8 & \textit{21.1} & 18.7 & 1.7 & 23.5 & \textbf{26.5} 
\\ \hline
$\text{ConSE}^{}$  & \textit{8.3} & 19.5 & 21.6 & 18.0 & 21.1 & \textit{14.1} & 15.1 & 14.9 & 16.8 & 18.4 & \textit{14.0} & 17.7 & 19.9 & 17.6 & \textbf{23.4} 
\\ \hline
$\text{Devise}^{}$  & \textit{12.6} & 17.0 & 15.8 & 19.0 & 19.2 & \textit{14.6} & 16.3 & 9.9 & 18.4 & 14.8 & \textit{16.0} & 13.2 & 13.7 & 17.4 & \textbf{22.5} 
\\
\hline
$\text{SynC}_{o-vs-o}^{}$ & \textit{15.3} & 19.8 & 17.3 & 20.3 & 21.3 & \textit{17.6} & 17.2 & 17.6 & 21.6 & 20.5 & \textit{17.0} & 15.0 & 15.7 & 20.2 & \textbf{24.0} 
\\ \hline

\multicolumn{16}{c|}{Awa2 dataset} \\
\hline
$\text{Linear}_{V \rightarrow S}^{}$ & \textit{31.1} & 40.2 & 38.5 & \textbf{43.6} & 37.9 & \textit{40.4} & 26.9 & 34.6 & 40.5 & 43.3 & \textit{42.1} & 39.9 & 28.1 & 38.5 & 41.6 
\\ \hline
$\text{Linear}_{S \rightarrow V}^{}$ & \textit{38.1} & 44.1 & 49.7 & 53.9 & 55.0 & \textit{56.6} & 42.4 & 48.1 & 41.2 & \textbf{57.7}  & \textit{54.7} & 49.3 & 14.4 & 50.4 & 46.5 
\\ \hline
ESZSL   & \textit{40.9} &  42.2 & 55.8 & 53.1 & 57.1 & \textit{\textbf{61.4}} & 37.7 & 49.0 & 48.2 & 44.3  & \textit{48.2} & 37.6 & 7.9 & 49.7 & 54.6 
\\ \hline
$\text{ConSE}^{}$  & \textit{27.4} & 31.3 & 34.3 & \textbf{43.3} & 39.2 & \textit{31.3} & 27.4 & 29.8 & 38.4 & 41.4 & \textit{34.7} & 31.3 & 16.7 & 42.3 & 42.1 
\\ \hline
$\text{Devise}^{}$  & \textit{37.2} & 34.1 & 46.6 & 33.7 & 43.4 & \textit{43.2} & 42.6 & 44.9 & 30.6 & 36.4 & \textit{\textbf{52.0}} & 40.7 & 13.5 & 32.7 & 37.6 
\\ \hline
$\text{SynC}_{o-vs-o}^{}$   & \textit{43.9} & 41.1 & 45.8 & 47.1 & 47.5 & \textit{46.9} & 46.6 & 47.4 & 50.0 & 52.1 & \textit{\textbf{53.3}} & 40.0 & 15.2 & 45.5 & 48.1 
\\ \hline
\end{tabular}
}
\end{center}
\end{table*}

\subsection{Comparison to other approaches}

The main results of the evaluation are reported in Table~\ref{tab:imagenet} for ImageNet. 
They confirm the relevance of our method and text collections to learn semantic prototypes for ZSL, as the best results are consistently obtained with our prototypes. Specifically, for ImageNet, the best result reported on the unbiased split in \cite{hascoet19cvpr} is 14.1 with ADGPM, and 13.5 with a ``traditional'' ZSL model (not making use of additional nodes in the class hierarchy), which used GloVe embeddings pretrained on Common Crawl. By contrast, our best result is
17.2 with FastText, obtained with embeddings trained on a much smaller dataset.
We also provide results for CUB and AwA2 in Table~\ref{tab:cub_awa2}. These results are less relevant since manual attributes exist for these smaller scale datasets, but still provide interesting insights. Importantly, these results are obtained using \emph{unsupervised} prototypes and should not be directly compared to results obtained with manual attributes.
On CUB, the best results are obtained with the embeddings learned on the $fl_{cust}$ collection for the three configurations and significantly outperform previous embeddings.
Interestingly, there does not seem to be a clear tendancy on AwA2. It turns out that performance obtainable with unsupervised prototypes on AwA2 is already quite close to performance with manual attributes -- see Sec.~\ref{sec:comparison_manual_attributes}. Our method is therefore unable to provide a significant improvement, unlike on the other two datasets.

Within each embeddings methods for all three datasets, the best results are usually obtained with $fl_{cust}$ and $fl_{wiki}$ usually performs better than baseline methods. 
The gain is especially large when compared to the largest available pretrained models for word2vec and FastText. 
This result is obtained although the largest collections used to create pretrained embeddings are 2 to 3 orders of magnitude larger than the collections we use.
For GloVe on ImageNet, the model pretrained on Common Crawl has the best performance. This embedding has poor behavior for all smaller scale datasets, indicating that the combination of local and global contexts at its core is able to capture interesting information at large scale.
While its performance on the smaller pretrained dataset is significantly lower than that of FastText, the two models are nearly equivalent when trained on Common Crawl. 
A similar finding was reported for text classification tasks~\cite{mikolov-etal-2018-advances}.
The strong performance of $fl_{cust}$ follows intuition since the collection was specifically built to cover the concepts which appear in the three test dataset.
This finding confirms the usefulness of smaller but adapted collections for NLP applications such as medical entity recognition~\cite{el-boukkouri-etal-2019-embedding} or sentiment analysis~\cite{kameswara-sarma-etal-2018-domain}.
Note that we also combined $fl_{wiki}$ and $fl_{cust}$ to obtain a more generic Flickr model. The obtained results were only marginally better compared to the single use of $fl_{cust}$ and are reported in the supplementary material.

We note that FastText and word2vec are better than GloVe for ImageNet and CUB, the two largest datasets with respect to number of classes.
Overall, the best performance is usually obtained with $fl_{cust}$ and FastText embeddings.

\subsection{Influence of text collection size}\label{sec:text_size}

\begin{table}[tb]
\caption{ZSL performance with 0\%, 50\%, 75\% and 90\% data removed from wiki and $fl_{cust}$ collections, on the ImageNet dataset. With FastText embeddings.}
\label{tab:collection_size}
\begin{center}
\begin{tabular}{|l|l|l|l|l|l|}
 \hline
Collection & Data removed   & 0\% & 50\% & 75\% & 90\%  
\\ \cline{1-6}

\multirow{4}{*}{wiki}
& $\text{Linear}_{S \rightarrow V}$  & 12.1 & 11.6 & 11.3 & 10.2 
\\ \cline{2-6} 
 & ESZSL    & 10.1 &  9.8 & 9.9 & 9.6
\\ \cline{2-6} 
& $\text{ConSE}$    & 10.5 &  11.0 & 10.5 & 9.9
\\ \cline{2-6} 
& $\text{Devise}$    & 10.1 &  8.3 & 8.7 & 8.0
\\ \hline 

\multirow{4}{*}{$fl_{cust}$}
 & $\text{Linear}_{S \rightarrow V}$  & 17.2 & 16.8 & 16.3 & 15.6
\\ \cline{2-6} 
 & ESZSL    & 15.8 &  15.1 & 15.3 & 14.3
\\ \cline{2-6} 
& $\text{ConSE}$    & 14.5 &  14.1 & 14.1 & 14.3
\\ \cline{2-6} 
& $\text{Devise}$    & 13.8 &  13.4 & 13.2 & 12.5
\\ \hline 

\end{tabular}
\end{center}
\end{table}

The quality of semantic embeddings is influenced by the size of the text collections used to learn them.
Existing comparisons are usually done among different collections~\cite{mikolov2013w2v,pennington-etal-2014-glove,bojanowski-etal-2017-enriching}.
While interesting, these comparisons do not provide direct information about the robustness of each collection.
To test robustness, we ablate 50\%, 75\% and 90\% of $fl_{cust}$ and wiki collections and report results for ImageNet using FastText embeddings in Table~\ref{tab:collection_size}.
Performance is as expected correlated to the collection size, with the best results being obtained for full text collections and the worst when 90\% of them is removed.
Interestingly, the performance drop is not drastic for either of the collection. 
For instance, with only 10\% of the initial collections, accuracy drops from 12.5 to 10.4 for $wiki$ (16.8\% relative drop) and from 17.7 to 15.6 for $fl_{cust}$ (11.8\% relative drop). 
The relative drop is smaller for $fl_{cust}$ compared to wiki; 
this indicates that a collection which is adapted for the task is more robust to changes in the quantity of available data.

\subsection{Comparison to manual attributes} \label{sec:comparison_manual_attributes}

\begin{figure}
    \centering
  \includegraphics[width=12cm]{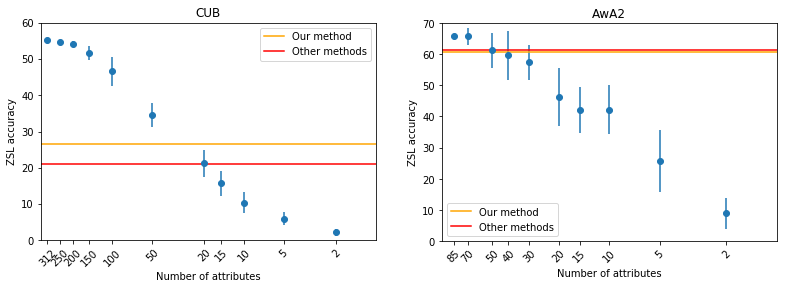}
    \caption{Ablation of manual attributes on CUB and AwA2. Measured with the linear model, averaged over 10 runs. Best results for prototypes based on word embeddings are also reported.}
    \label{fig:cub_amputation}
\end{figure}

Although our webly semantic prototypes enable to achieve much better results than with previously available prototypes extracted from text corpora, it is still interesting to compare them to what can be achieved with hand-crafted attributes. Such attributes do not exist for very large scale datasets such as ImageNet, but they are provided with smaller scale datasets such as CUB and AwA2.

To quantify how much better hand-crafted prototypes perform when compared to webly supervised prototypes, we conducted an ablation study on CUB attributes similar to Sec.~\ref{sec:text_size}. We started with the full list of attributes, initially comprising 312 attributes for each bird species, and randomly removed attributes while measuring the resulting ZSL score. The scores where obtained with the $\text{Linear}_{S \rightarrow V}$ model due to its good results, robustness and simplicity. To account for the noise caused by to the randomness of the removed attributes, each reported score is the average of 10 measurements, each with different random attributes removed. The remaining attributes are $\ell 2$-normalized, and the hyper-parameter is re-selected by cross-validation for each run.
Fig.~\ref{fig:cub_amputation} provides a visualization of the result; a table with the exact scores is available in the supplementary materials.

On CUB, there is still a substantial margin for improvement; even though our method enables a significant increase over other methods, it is still barely above results achievable by selecting only 20 attributes among the 312 initial attributes.
Interestingly, the difference between webly supervised and hand-crafted prototypes is not so pronounced on the AwA2 dataset; the ZSL accuracy between the two settings is even surprisingly close. This may be explained by the fact that AwA2 only contains 10 test classes; class prototypes need not enable a ZSL model to subtly distinguish very similar classes. Consequently, our best result is comparable to the best result enabled by previous methods.

\subsection{Error analysis}

\begin{figure}[tb]
    \centering
    \includegraphics[width=12cm]{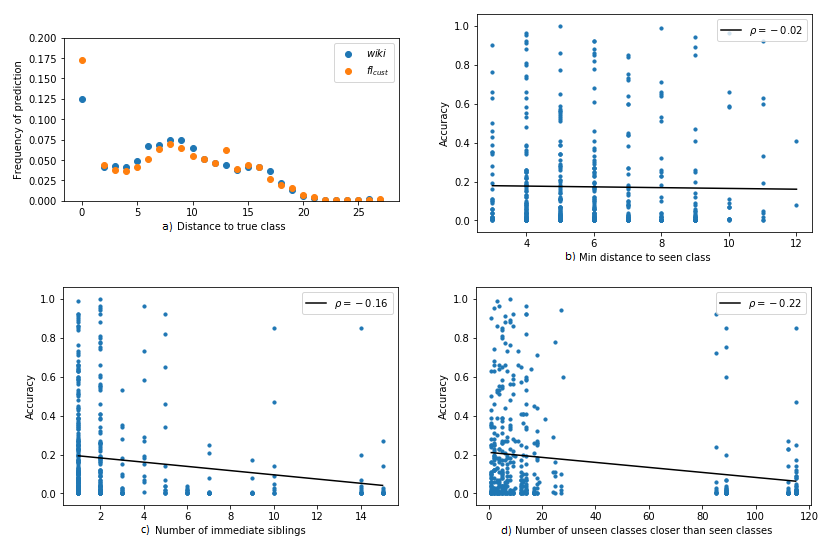}
    \label{fig:distance_and_siblings_correlation}
    \caption{
    (a) Distance from predicted class to correct class in the WordNet hierarchy.
    Correlation $\rho$ between ZSL accuracy and
    (b) distance to the closest seen class
    (c) the number of immediate unseen test class siblings
    (d) the number of unseen classes closer than the closest seen class,
    for all 500 unseen ImageNet classes.
    \label{fig:correlations}
    }
\end{figure}

We analyze how far incorrect predictions are from the correct class by computing the distance between the predicted class and the correct class.
We define the distance between two classes as the shortest path between them in the WordNet hierarchy.
For a given distance $d$, we measure the number of predictions that are exactly $d$ nodes away from the correct class -- a distance of 0 being a correct prediction.
Results for $wiki$ and $fl_{cust}$ are presented in Figure~\ref{fig:correlations}(a); the general tendency seems to be that classes farther away from the correct class are less likely to be predicted.
Note that no two test classes are a distance of one from each other, since it is not possible for a test class to be a direct parent or child of another test class.

We further analyze the main factors behind classification errors. Experiments below are conducted on ImageNet, with the $\text{Linear}_{S \rightarrow V}$ model trained using the FastText $fl_{cust}$ embeddings.
Our first hypothesis was that the distance between unseen and seen classes influences classification accuracy: the less an unseen class resembles any seen class, the harder it is to identify. 
To test this hypothesis, we consider for each unseen class $c_u$ the minimal distance to a seen class $\underset{c \in \mathcal{C}_s}{\text{min}}~d(c_u, c)$, and analyze its relation to the prediction accuracy.
The resulting plot is displayed in Figure~\ref{fig:correlations}(b). Surprisingly, the distance to the closest seen class seems to have little to no effect on the accuracy (correlation $\rho=-0.02$).

Another hypothesis was that unseen classes close to other unseen classes are harder to classify than isolated unseen classes, as more confusions are possible. For each unseen class, we therefore compute the number of immediate siblings, a sibling being defined as an unseen class having the same parent in the WordNet hierarchy as the reference (unseen) class. The link between this metric and class accuracy is slightly stronger, with a correlation $\rho = -0.16$ as illustrated in Figure~\ref{fig:correlations}(c), but still weak overall.

\begin{figure}[th]
    \centering
    \includegraphics[width=12cm]{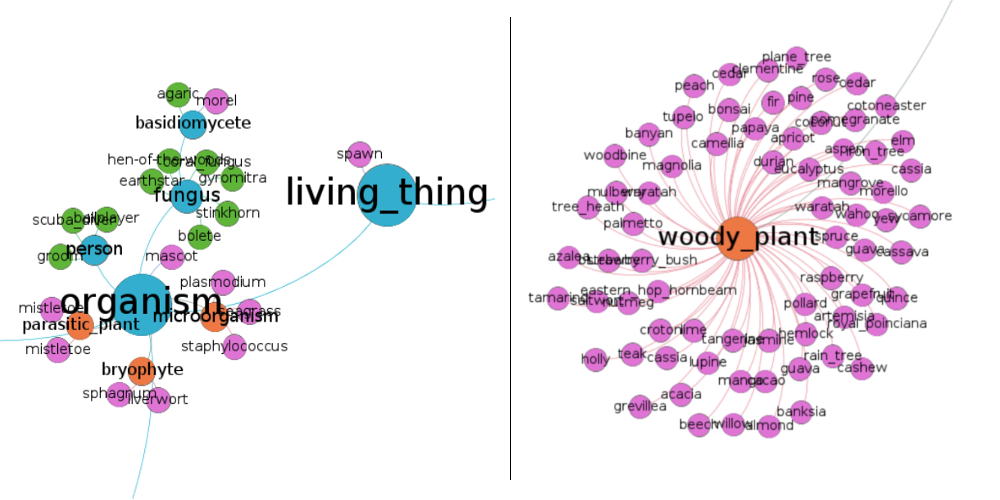}
    \caption{Graph visualization of parts of the WordNet hierarchy. Green and pink leaves are resp. seen and unseen classes. Intermediate nodes are orange if there is no seen class among their children, and blue otherwise. Full graph is available in the supp. materials.}
    \label{fig:graph}
\end{figure}

We combine these two hypotheses by considering the number of unseen classes closer than the closest seen class for each unseen class. The link with class accuracy is more pronounced than by simply considering the number of siblings, with a correlation $\rho = -0.22$ as illustrated in Fig.~\ref{fig:correlations}(d).
Examples of classes at both ends of the spectrum are visible in Figure~\ref{fig:graph}: unseen class \textit{morel} (on the left) is close to seen class \textit{agaric} and has no unseen siblings; its class accuracy is $0.63$. On the other hand, classes \textit{holly}, \textit{teak} and \textit{grevillea} (on the right) have many unseen siblings and are far from any seen class; their respective accuracy are $0.01$, $0.00$ and $0.03$. More generally, classes which are descendant of the intermediate node \textit{woody plant} have an average accuracy of $0.053$.
The full graph visualization of the 1000 training classes, 500 testing classes and intermediate nodes of the ImageNet ZSL dataset is provided in the supplementary materials.

\section{Conclusion}

We proposed a new method to build semantic class prototypes automatically, thus enabling to better address large scale ZSL.
Our results indicate that appropriately learning embeddings on specialized collections made of photo metadata is better than exploiting generic embeddings as it was done previously in ZSL.
This still stands when generic embeddings are learned with collections which are two to three orders of magnitude larger than specialized collections.
Among photo metadata based collection, the use of Flickr seems preferable to that of metadata associated to photos from Web pages. 
This is notably an effect of a better semantic coverage of classes in Flickr compared to ClueWeb12.
We will release the specialized embeddings created here, as well as the visual features and the code necessary to produce the reported results in order to facilitate reproducibility.



\bibliographystyle{splncs04}
\bibliography{main}

\appendix

\title{Supplementary Material to: Webly Supervised Semantic Embeddings for Large Scale Zero-Shot Learning}
\titlerunning{Supplementary Material: Webly Supervised Semantic Embeddings}
\authorrunning{Y. Le Cacheux, A. Popescu and H. Le Borgne}

\author{}
\institute{}

\maketitle


\section{Implementation details}

\subsection{Word embeddings}

We provide additional details about the versions of the embedding implementations used, namely word2vec, GloVe and FastText, as well as their parameters.

We used the original implementation of each method available at:
\begin{itemize}
    \item word2vec - \url{https://code.google.com/archive/p/word2vec/}
    \item GloVe - \url{https://nlp.stanford.edu/software/GloVe-1.2.zip}
    \item{FastText} - \url{https://github.com/facebookresearch/fastText}
\end{itemize}

The main parameters used for to create semantic embeddings are given in Table~\ref{tab:params}. These values were selected by following the guidelines from the original papers.
We ran initial tests with a larger number of epochs and this did not improve results compared to the numbers presented in Table~\ref{tab:params}.

\begin{table}
\caption{Training parameters for the different semantic embedding models.}
\label{tab:params}
\begin{center}
\begin{tabular}{|l|l|l|l|}
\hline
Parameter & word2vec & GloVe & FastText 
\\ \hline
Epochs & 25 & 100 & 25 
\\ \hline
Learning rate & 0.1 & 0.05 & 0.1 
\\ \hline
Window & 10 & 10 & 10 
\\ \hline
Embedding dimension & 300 & 300 & 300 
\\ \hline

\end{tabular}
\end{center}

\end{table}

The set of parameters used each time in order to facilitate reproducibility is reported in Table~\ref{tab:more_params}.
We exclude the input, output and intermediary, as well as the number of threads because they do not influence directly the learning process.

\begin{table*}
\caption{Command line used to train embeddings.}
\label{tab:more_params}
\begin{center}
\begin{tabular}{|l|l|}
\hline
\textbf{Model} & \textbf{Command} 
\\ \hline
word2vec &  -size 300 -window 1 -sample 1e-4 -negative 5 -hs 0 -binary 0 \\
         & -cbow 0 -iter 25 -min-count 5
\\ \hline
GloVe & -x-max 100 -iter 100 -eta 0.05 -vector-size 300 -alpha 0.75
\\ \hline
FastText & skipgram -dim 300 -epoch 25 -minn 4 -maxn 6 -lr 0.1 -ws 10 -minCount 5
\\ \hline

\end{tabular}
\end{center}
\end{table*}

We tried to add phrase representations \cite{mikolov2013w2v}, 
but it did not provide any improvement of results in ZSL experiments, thus it was not used in the final models.


\subsection{Visual features and ZSL models}

For the ImageNet dataset, we use visual features provided by Hascoet \textit{et al.} \cite{hascoet19cvpr}, 
which consist in the weights of the last pooling layer of a pre-trained ResNet. We also use a pre-trained ResNet to extract visual features for the CUB and AwA2 datasets, and we further apply 10-crop to the images.

On ImageNet and CUB, hyper-parameters of ZSL methods are selected using respectively 200 and 50 random classes for validation. For AwA2, we use the 8 classes which are not in the ILSVRC out of the 40 training classes.

\subsection{Datasets}

Some statistics regarding the word word frequencies in each dataset are available in Table~\ref{tab:word_frequency}.

\begin{table}
\caption{Mean word frequency and standard deviation (in thousands of occurrences) in a corpus for words present in a given dataset.}
\label{tab:word_frequency}
\begin{center}
  {
\begin{tabular}{|l|l|l|l|l|}
\hline
 & $wiki$ & $clue$ & $fl_{wiki}$ & $fl_{cust}$ 
\\ \hline
ImageNet & $51 \pm 192$ & $183 \pm 886 $ & $ 49 \pm 150.2 $  & $ 56 \pm 149.7 $  
\\ \hline
CUB & $104 \pm 275$ & $416 \pm 1596$ & $117.9 \pm 260.6$  & $ 146 \pm 303.8 $ 
\\ \hline
AwA2 & $40 \pm 94$ & $320 \pm 714$ & $73.1 \pm 160.3$ & $116.8 \pm 207.9$
\\ \hline
\end{tabular}
}
\end{center}
\end{table}


\section{Additional results}


We provide results for $\text{Linear}_{V \rightarrow S}$, $\text{Linear}_{S \rightarrow V}$ with no $\ell$2 normalization applied to attributes, as well as for ESZSL with normalization as we found that normalizing attributes could have a significant impact on these models.
Results are provided for ImageNet (Table~\ref{tab:imagenet_supp}) as well as CUB and AwA2 (Table~\ref{tab:cub_awa2_supp}) similarly to tables \ref{tab:imagenet} and \ref{tab:cub_awa2} of the main paper.

\begin{table*}[tb]
\caption{Results with and without $\ell$2 normalization of attributes on the ImageNet dataset; this table is similar to Table~\ref{tab:imagenet} of the main paper. Normalized attributes are indicated with the $norm$ exponent; results without the exponent correspond to unnormalized attributes.
}
\label{tab:imagenet_supp}
\begin{center}
\resizebox{\textwidth}{!}{
\begin{tabular}{|l|l|l|l|l|l|l|l|l|l|l|l|l|l|l|l|}
\hline
Model &  \multicolumn{5}{c|}{word2vec} &  \multicolumn{5}{c|}{GloVe}&  \multicolumn{5}{c|}{FastText}
\\ \hline
Source & \textit{pt} & wiki & clue & $\mathbf{fl_{wiki}}$  & $\mathbf{fl_{cust}}$ & \textit{pt} & wiki & clue & $\mathbf{fl_{wiki}}$  & $\mathbf{fl_{cust}}$ & \textit{pt} & wiki & clue & $\mathbf{fl_{wiki}}$  & $\mathbf{fl_{cust}}$
\\ \hline
$\text{Linear}_{V \rightarrow S}$ & \textit{2.0} & 4.3 & 4.1 & 3.7 & 4.6 & \textit{5.4} & 3.3 & 1.3 & 4.2 & 5.3 & \textit{1.8} & 4.6 & 1.1 & 4.0 & \textbf{4.9}
\\ \hline
$\text{Linear}_{S \rightarrow V}$ & \textit{10.7} & 12.1 &  12.5 & 12.4 & 17.0 & \textit{14.3} & 7.7 & 8.7 & 8.2 & 10.6 & \textit{14.6} & 12.5 & 2.5  & 12.8 & \textbf{17.3} 
\\ \hline
$\text{ESZSL}^{norm}$ & \textit{13.4} & 12.8 &  13.6 & 13.8 & \textbf{18.0} & \textit{16.1} & 10.7 & 11.9 & 13.7 & 14.4  & \textit{ 16.0} & 13.0 & 8.6  & 14.7 & 17.7 
\\ \hline

\end{tabular}
}
\end{center}
\end{table*}

\begin{table*}[bt]
\caption{
Results with and without $\ell$2 normalization of attributes on the CUB and AwA2 datasets; this table is similar to Table~\ref{tab:cub_awa2} of the main paper. Normalized attributes are indicated with the $norm$ exponent; results without the exponent correspond to unnormalized attributes.
}
\label{tab:cub_awa2_supp}
\begin{center}
\resizebox{\textwidth}{!}{
\begin{tabular}{|l|l|l|l|l|l|l|l|l|l|l|l|l|l|l|l|}
\hline
Model &  \multicolumn{5}{c|}{word2vec} &  \multicolumn{5}{c|}{GloVe}&  \multicolumn{5}{c|}{FastText}
\\ \hline
Source & \textit{pt} & wiki & clue & $\mathbf{fl_{wiki}}$  & $\mathbf{fl_{cust}}$ & \textit{pt } & wiki & clue & $\mathbf{fl_{wiki}}$  & $\mathbf{fl_{cust}}$ & \textit{pt} & wiki & clue & $\mathbf{fl_{wiki}}$  & $\mathbf{fl_{cust}}$
\\ \hline
\multicolumn{16}{|c|}{CUB dataset} \\
\hline
$\text{Linear}_{V \rightarrow S}$ & \textit{5.6} & 12.1 & 10.7 & 11.7 & \textbf{15.7} & \textit{3.9} & 13.4 & 5.5 & 11.5 & 12.1 & \textit{3.2} & 12.0 & 7.9 & 11.7 & 15.2
\\ \hline
$\text{Linear}_{S \rightarrow V}$ & \textit{14.3} &  19.0 & 17.7 & 20.1 & 21.3 & \textit{20.6} & 14.3 & 12.9 & 14.9 & 17.3 & \textit{18.0} & 17.4 & 2.0  & 19.2 & \textbf{22.4}
\\ \hline 
$\text{ESZSL}^{norm}$  & \textit{16.9} & 20.6 & 16.7 &  20.9 & 23.6 & \textit{19.1} & 18.3 & 18.8 & 21.2 & 22.0 & \textit{20.7} & 17.4 & 19.9 & 21.5 & \textbf{24.0} 
\\ \hline


\multicolumn{16}{|c|}{Awa2 dataset} \\
\hline
$\text{Linear}_{V \rightarrow S}$ & \textit{27.3} & 15.6 & 33.6 & 15.5 & 25.9 & \textit{30.6} & 17.2 & 34.9 & 26.3 & \textbf{42.3} & \textit{7.8} & 11.8 & 9.7 & 3.8 & 15.2
\\ \hline
$\text{Linear}_{S \rightarrow V}$ & \textit{24.8} &  45.0 & 53.2 & 56.1 & 56.6 & \textit{55.7} & 48.4 & 50.5 & 41.7 & \textbf{60.6}  & \textit{58.1} & 47.6 & 2.2 & 47.9 & 55.2 
\\ \hline
$\text{ESZSL}^{norm}$  & \textit{41.6} &  38.7 & 46.7 & 49.5 & 45.6 & \textit{55.3} & 31.6 & 47.0 & 48.5 & 46.4 & \textit{\textbf{55.9}} & 38.2 & 18.6 & 45.3 & 43.9 
\\ \hline

\end{tabular}
}
\end{center}
\end{table*}

For comparison with other papers, we also provide top-5 and top-10 accuracy for the $\text{Linear}_{S \rightarrow V}^{norm}$ model trained on FastText $fl_{cust}$ in Table~\ref{tab:topn}.

\begin{table}
\caption{Top-k accuracy on ImageNet, with FastText and $fl_{cust}$.}
\label{tab:top5}
\begin{center}
\begin{tabular}{|l|l|l|l|}
\hline
 & top-1 & top-5 & top-10 
\\ \hline
$\text{Linear}_{S \rightarrow V}$ & 17.3 & 39.6 & 51.9 
\\ \hline
$\text{Linear}_{S \rightarrow V}^{norm}$ & 17.2 & 39.2 & 51.4 
\\ \hline
$\text{ESZSL}$ & 15.8 & 37.5 & 49.3 
\\ \hline
$\text{ESZSL}^{norm}$ & 17.7 & 40.0 & 51.4 
\\ \hline
$\text{ConSE}^{norm}$ & 14.5 & 32.4 & 42.0 
\\ \hline
$\text{Devise}^{norm}$ & 13.8 & 32.1 & 43.7 
\\ \hline
\end{tabular}
\end{center}
\label{tab:topn}
\end{table}

\section{Effect of User Voting on Flickr Embeddings}

In Section~3 of the main article, we reported the introduction of user voting instead of raw co-occurrence frequency in Flickr in order the quality of embeddings. 
When user voting is exploited, each user gets to vote only once for a pair of words and the effect of bulk tagging is thus reduced.
We compare the $fl_{cust}$ results presented in Table~1 of the main paper, obtained with user voting and those of $fl_{cust}^{raw}$, obtained with a simple count of word co-occurrences. 
We use FastText and all the tested ZSL methods of the main paper.  
The results, presented in Table~\ref{tab:voting}, confirm that user voting has a positive effect for all collection sizes and ZSL methods tested.
This confirms the importance of an appropriate preprocessing of text collections.

\begin{table}[]
\caption{ZSL accuracy on the ImageNet dataset for two versions of the $fl_{cust}$ collection which exploit user voting ($fl_{cust}$) and raw counts ($fl_{cust}^{raw}$) to compute word co-occurrences.
}
\begin{center}
\begin{tabular}{|l|l|l|}
\hline
Model &  \multicolumn{2}{c|}{FastText} 
\\ \hline
Source & $fl_{cust}$ & $fl_{cust}^{raw}$
\\ \hline
$\text{Linear}_{S \rightarrow V}$ & 17.3 & 13.9 
\\ \hline
$\text{Linear}^{norm}_{S \rightarrow V}$ & 17.2 &  13.8
\\ \hline
ESZSL & 15.8 &  12.5
\\ \hline
$\text{ESZSL}^{norm}$ & 17.7 & 15.5
\\ \hline
$\text{ConSE}^{norm}$  & 14.5 & 12.6
\\ \hline
$\text{Devise}^{norm}$  & 13.8 & 11.2
\\ \hline
\end{tabular}
\end{center}
\label{tab:voting}
\end{table}


\section{Effect of combining $fl_{cust}$ and $fl_{wiki}$}
In Subsection 4.2 of the main paper, we noted that $fl_{cust}$, the Flickr collection which includes metadata from the three test datasets, gave the best results among the text collections tested. 
Since $fl_{wiki}$ is collected from the same source but with a different set of concepts, we merged the two collections to observe the effect of results.
The results are reported in Table~\ref{tab:merged} and they confirm that most of the performance gain is due to the use of $fl_{cust}$.

\begin{table*}[]
\caption{ZSL accuracy for the ImageNet dataset.}
\begin{center}
\begin{tabular}{|l|l|l|l|l|l|l|l|l|l|}
\hline
Model &  \multicolumn{3}{c|}{word2vec} &  \multicolumn{3}{c|}{GloVe}&  \multicolumn{3}{c|}{FastText}
\\ \hline
Source & $fl_{wiki}$  & $fl_{cust}$ & $fl_{merged}$ & $fl_{wiki}$  & $fl_{cust}$ & $fl_{merged}$ & $fl_{wiki}$  & $fl_{cust}$ & $fl_{merged}$
\\ \hline
$\text{Linear}_{S \rightarrow V}$ & 12.4 & 17.0 & 17.2 & 8.2 & 10.6 & 11.1 & 12.8 & 17.3 & 17.2
\\ \hline
$\text{Linear}^{norm}_{S \rightarrow V}$ & 12.8 & 17.1 & 16.9 & 9.2 & 11.4 & 11.9 & 13.3 & 17.2 & 17.1
\\ \hline
ESZSL & 9.5 & 15.3 & 15.3 & 11.1 & 12.0 & 14.4 & 11.9 & 15.8 & 15.2
\\ \hline
$\text{ESZSL}^{norm}$ & 13.8 & 18.0 & 17.9 & 13.7 & 14.4 & 17.1 & 14.7 & 17.7 & 17.9
\\ \hline
$\text{ConSE}^{norm}$  & 11.9 & 13.5 & 14.1 & 11.3 & 11.9 & 12.7 & 12.6 & 14.5 & 14.2
\\ \hline
$\text{Devise}^{norm}$  & 9.6 & 13.3 & 13.9 & 3.8 & 3.4 & 9.0 & 10.3 & 13.8 & 13.6
\\ \hline
\end{tabular}
\end{center}
\label{tab:merged}
\end{table*}


\section{Comparison with manual attributes}

Table~\ref{tab:attributes_amputation} contains the data used to create Figure~1 of the main paper. Note that when all attributes are selected, there is no randomness involved since $\text{Linear}^{}_{S \rightarrow V}$ is deterministic, hence a standard deviation of $0$.

\begin{table}[tb]
\caption{Performance with linear model on CUB and AwA2 with attributes randomly removed. Averaged on 10 runs.}
\label{tab:attributes_amputation}
\begin{center}
\begin{tabular}{|l|c|l|l|l|l|l|l|l|l|l|l|l|l|}
\hline
\multicolumn{12}{|c|}{CUB} \\
\hline
Number of attributes & 312 & 250 & 200 & 150 & 100 & 50 & 20 & 15 & 10 & 5 & 2
\\ \hline
Mean ZSL score & 55.3 & 54.8 & 54.2 & 51.7 & 46.6 & 34.7 & 21.2 & 15.7 & 10.4 & 5.9 & 2.2
\\ \hline
Standard deviation & 0.0 & 0.5 & 0.9 & 1.9 & 3.9 & 3.3 & 3.8 & 3.5 & 2.9 & 1.8 & 0.9
\\ \hline
\multicolumn{12}{|c|}{AwA2} \\
\hline
\multicolumn{2}{|l|}{Number of attributes} & 85 & 70 & 50 & 40 & 30 & 20 & 15 & 10 & 5 & 2
\\ \hline
\multicolumn{2}{|l|}{Mean ZSL score} & 66.0 & 65.8 & 61.3 & 59.7 & 57.4 & 46.2 & 42.2 & 42.2 & 25.7 & 8.8
\\ \hline
\multicolumn{2}{|l|}{Standard deviation} & 0.0 & 2.8 & 5.7 & 7.9 & 5.6 & 9.3 & 7.4 & 7.9 & 10.1 & 4.6
\\ \hline

\end{tabular}
\end{center}
\end{table}

\section{Performance gain of $fl_{cust}$ over $wiki$}

\begin{figure*}[t!]
    \centering
    \begin{subfigure}[t]{0.5\textwidth}
        \centering
        \includegraphics[width=6cm, height=4cm]{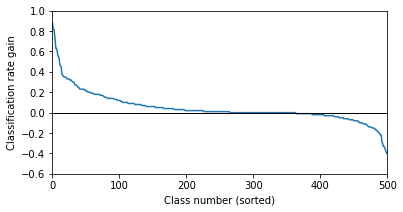}
        \caption{$\text{Linear}_{S \rightarrow V}$ model}
        \label{fig:performance_gain1}
    \end{subfigure}%
    ~ 
    \begin{subfigure}[t]{0.5\textwidth}
        \centering
        \includegraphics[width=6cm]{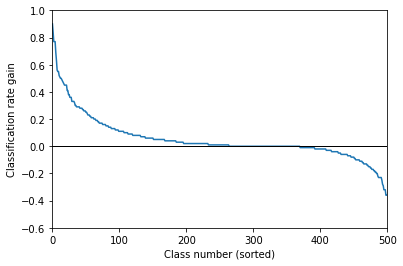}
        \caption{$\text{Linear}^{norm}_{S \rightarrow V}$} \label{fig:performance_gain2}
    \end{subfigure}%
 
    \begin{subfigure}[t]{0.5\textwidth}
        \centering
        \includegraphics[width=6cm]{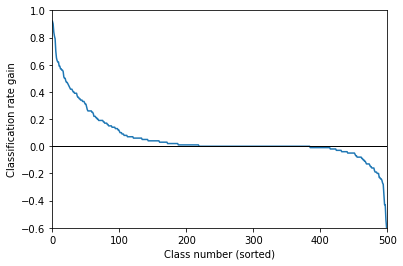}
        \caption{$\text{ESZSL}$}\label{fig:performance_gain3}
    \end{subfigure}%
    ~ 
    \begin{subfigure}[t]{0.5\textwidth}
        \centering
        \includegraphics[width=6cm]{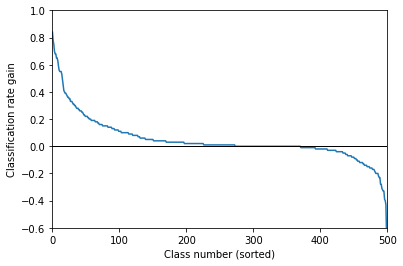}
        \caption{$\text{ESZSL}^{norm}$.}\label{fig:performance_gain4}
    \end{subfigure}
    
    \begin{subfigure}[t]{0.5\textwidth}
        \centering
        \includegraphics[width=6cm]{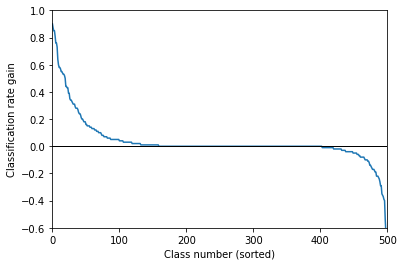}
        \caption{$\text{ConSE}^{norm}$}\label{fig:performance_gain5}
    \end{subfigure}%
    ~ 
    \begin{subfigure}[t]{0.5\textwidth}
        \centering
        \includegraphics[width=6cm]{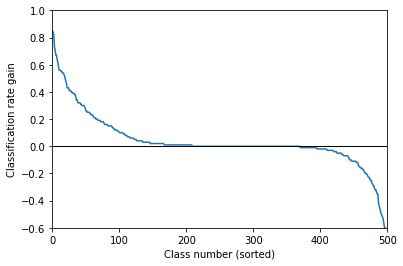}
        \caption{$\text{Devise}^{norm}$.}\label{fig:performance_gain6}
    \end{subfigure}
    \caption{Performance gain on each test class (by decreasing value) for the $fl_{cust}$ collection w.r.t $wiki$ collection, with several ZSL methods.
    }
\label{fig:gains}
\end{figure*}

We present a comparison of FastText accuracy obtained for $wiki$ and $fl_{cust}$ for the ImageNet dataset with different models.
Figure~\ref{fig:gains} provides a view of accuracy differences between $fl_{cust}$ and $wiki$ for ImageNet test classes.
These differences are plotted in decreasing order from left to right.
For the  $\text{Linear}_{S \rightarrow V}$ model, $fl_{cust}$ is better for 265 of ImageNet test classes, no change is observed for another 99 classes and $wiki$ provides better results for the remaining 136 classes.
For classes that perform better with $fl_{cust}$, the average gain is 0.13 and the maximal gain is 0.88. For those performing worse, the average loss is $-0.08$ and the maximal loss is $-0.4$. 
Trends are similar for other methods, indicating that performance gains are robust with respect to the ZSL methods used.

\section{ImageNet ZSL Full Graph}

\begin{figure*}
\begin{center}
\includegraphics[width=0.99\textwidth]{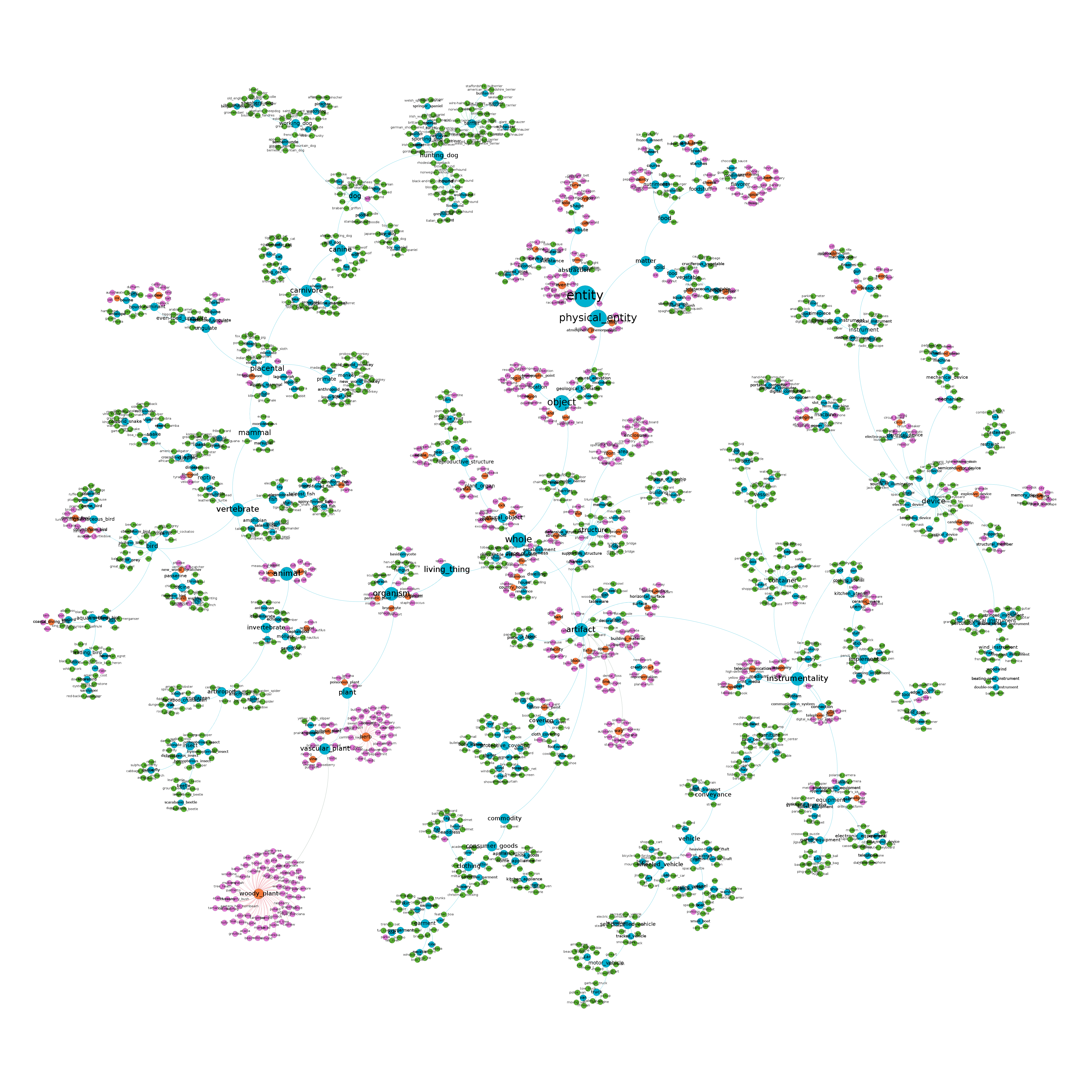}
\end{center}
\caption{Overview of the full class hierarchy. Pink nodes refer to test classes, green nodes refer to train classes, orange nodes have only test classes below them and blue nodes are other intermediate nodes. Best viewed in color with at least 600\% zoom.}
\label{fig:hierarchy}
\end{figure*}

We provide a visualization of the full WordNet hierarchy for all 1000 (resp. 500) training (resp. testing) classes, as well as some intermediate nodes in Fig.~\ref{fig:hierarchy}. We only keep one parent per node. Fig.~3 of the main paper contains subsets of this visualization. For nodes which originally have several hypernyms, we keep the nodes corresponding to the longest path to the root node \emph{``entity''}; we found that this leads to more meaningful paths, with fewer classes at each level. For example, we keep the path \emph{``greyhound''} $\rightarrow$ \emph{``hound''} $\rightarrow$ \emph{``hunting\_dog''} $\rightarrow$ \emph{``dog''} $\rightarrow$ \dots $\rightarrow$ \emph{``animal''} (visible in  Fig.~\ref{fig:hierarchy}) instead of \emph{``greyhound''} $\rightarrow$ \emph{``racer''} $\rightarrow$ \emph{``animal''}. We remove intermediate nodes which are not direct hypernyms of either a training or a testing class, as well as some other hand-picked nodes to improve readability.

In addition to the remarks from the main paper, it is interesting to observe that ZSL training and testing classes are not homogeneous in the hierarchy: some tree branches contain very few unseen classes, \emph{e.g.} \emph{``carnivore''}, while other contain many unseen classes and not a single seen class, \emph{e.g.} \emph{``woody\_plant''}. These latter classes appear very challenging to correctly predict.

\newpage

\end{document}